\documentclass[sigconf]{acmart}
\AtBeginDocument{%
  \providecommand\BibTeX{{%
    \normalfont B\kern-0.5em{\scshape i\kern-0.25em b}\kern-0.8em\TeX}}}

\copyrightyear{2021}
\acmYear{2021}
\setcopyright{rightsretained}
\acmConference[CIKM '21]{Proceedings of the 30th ACM International Conference on Information and Knowledge Management}{November 1--5, 2021}{Virtual Event, QLD, Australia}
\acmBooktitle{Proceedings of the 30th ACM International Conference on Information and Knowledge Management (CIKM '21), November 1--5, 2021, Virtual Event, QLD, Australia}
\acmDOI{10.1145/3459637.3481957}
\acmISBN{978-1-4503-8446-9/21/11}

\begin{document}

\title{TSI: an Ad Text Strength Indicator using Text-to-CTR and Semantic-Ad-Similarity}

\author{Shaunak Mishra}
\affiliation{%
  \institution{Yahoo Research}
   \country{USA}
}
\email{shaunakm@verizonmedia.com}\authornote{Equal contribution.}
\author{Changwei Hu}
\affiliation{%
  \institution{Yahoo Research}
   \country{USA}
}
\email{changweih@verizonmedia.com }
\authornotemark[1]

\author{Manisha Verma}
\affiliation{%
  \institution{Yahoo Research}
   \country{USA}
}
\email{manishav@verizonmedia.com}

\author{Kevin Yen}
\affiliation{%
  \institution{Yahoo Research}
   \country{USA}
}
\email{kevinyen@verizonmedia.com}

\author{Yifan Hu}
\affiliation{%
  \institution{Yahoo Research}
   \country{USA}
}
\email{yifanhu@verizonmedia.com}

\author{Maxim Sviridenko}
\affiliation{%
  \institution{Yahoo Research}
   \country{USA}
}
\email{sviri@verizonmedia.com}



\begin{abstract}
  Coming up with effective ad text is a time consuming process, and particularly challenging for small businesses with limited advertising experience. When an inexperienced advertiser onboards with a poorly written ad text, the ad platform has the opportunity to detect low performing ad text, and provide improvement suggestions. To realize this opportunity, we propose an ad text strength indicator (TSI) which: (i) predicts the click-through-rate (CTR) for an input ad text, (ii) fetches similar existing ads to create a neighborhood around the input ad, (iii) and compares the predicted CTRs in the neighborhood to declare whether the input ad is strong or weak. In addition, as suggestions for ad text improvement, TSI shows anonymized versions of superior ads (higher predicted CTR) in the neighborhood. For (i), we propose a BERT based text-to-CTR model trained on impressions and clicks associated with an ad text. For (ii), we propose a sentence-BERT based semantic-ad-similarity model trained using weak labels from ad campaign setup data.
  Offline experiments demonstrate that our BERT based text-to-CTR model achieves a significant lift in CTR prediction AUC for cold start (new) advertisers compared to bag-of-words based baselines.
  In addition, our semantic-textual-similarity model for similar ads retrieval achieves a precision@1 of $0.93$ (for retrieving ads from the same product category); this is significantly higher compared to unsupervised TF-IDF, word2vec, and sentence-BERT baselines.
  Finally, we share promising online results from advertisers in the Yahoo (Verizon Media) ad platform where a variant of TSI was implemented with sub-second end-to-end latency.
\end{abstract}


\begin{CCSXML}
<ccs2012>
<concept>
<concept_id>10002951.10003260.10003272</concept_id>
<concept_desc>Information systems~Online advertising</concept_desc>
<concept_significance>500</concept_significance>
</concept>
</ccs2012>
\end{CCSXML}
\ccsdesc[500]{Information systems~Online advertising}

\keywords{Online advertising; CTR prediction; semantic textual similarity; multilingual; ad creative; ad text optimization.}
%\keywords{datasets, neural networks, gaze detection, text tagging}

\maketitle

\section{Introduction} \label{sec:introduction}
\begin{figure}[]%!htb
\centering
  \includegraphics[width=1 \columnwidth]{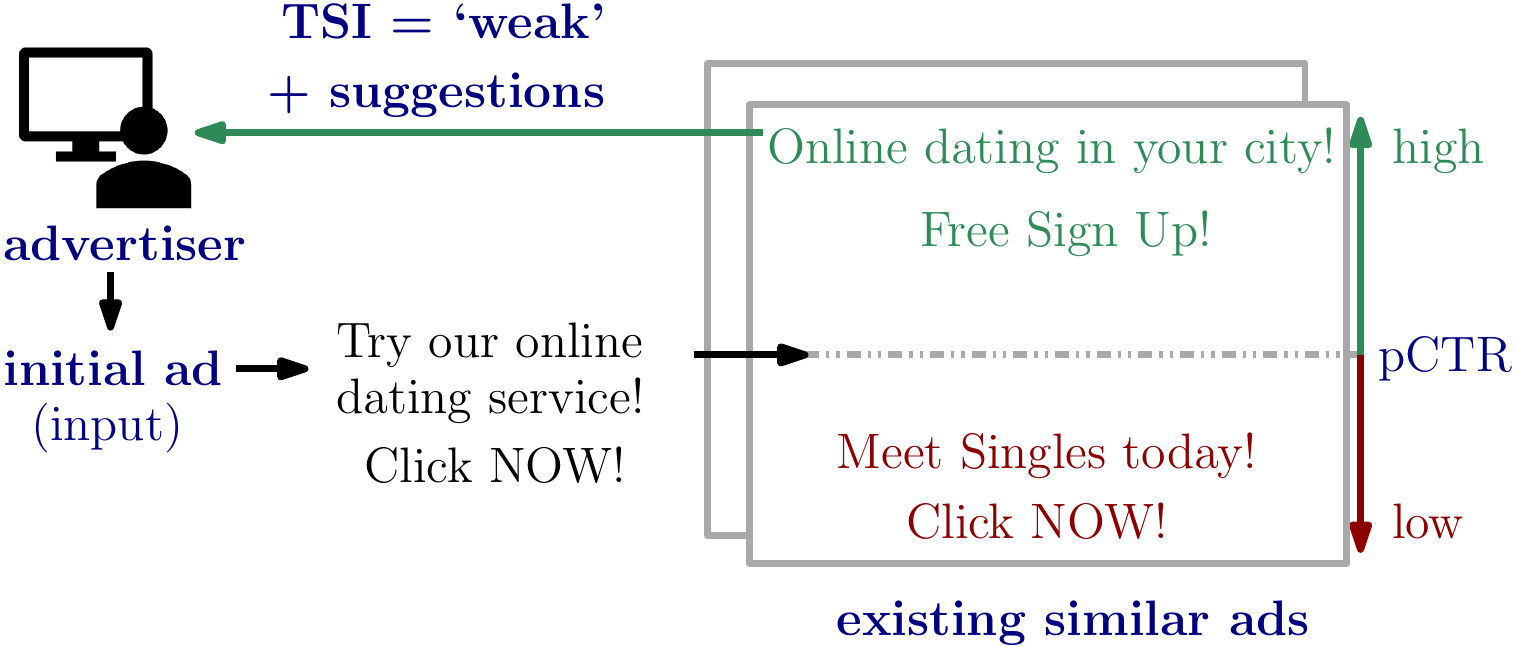}
  \caption{Illustrative example of a dating ad for which two similar ads already exist in the ad platform. One of the existing ads which highlights \textit{city} and \textit{free sign up} has higher predicted CTR (pCTR), and can be anonymized as a suggestion for the \textit{weaker} input ad.}
  \label{fig:pull_figure_main}
\end{figure}

Effective ad text can go a long way in attracting online users, and nudging them towards a purchase.
However, writing effective ad text is an inherently creative process, and may require years of experience. This can be challenging for small businesses, who have limited advertising budgets and may not be able to afford expert ad copywriters.
However, in an ad platform, there might be a mix of both well written and poorly written ads (as inferred from their online performance). Intuitively, it may be possible to help small businesses by comparing their ads with other semantically similar ads in the ad platform, and flagging ad text which is likely to perform relatively poor. In this paper, we build on this intuition, and introduce an ad text strength indicator (TSI).
As shown in Figure~\ref{fig:pull_figure_main}, the core idea behind TSI is to fetch semantically similar ads in an ad platform, compare their predicted click-through-rates (CTRs) as a proxy for ad-effectiveness, and flag the input ad as \textit{weak} if it has significantly stronger ads in its (semantic) neighborhood.
Furthermore, such relatively stronger ads can be anonymized (\emph{e.g.}, by removing brand references), and shown to the advertiser as suggestions for improvement (\emph{i.e.}, suggestions to inspire better ads and not to be copied verbatim).

%Trends \cite{taboola_trends}) to their design creatives.

%In addition, by improving ads across the board, TSI has the potent
%Higher eCPMs . better for user experience.

The TSI illustration in Figure~\ref{fig:pull_figure_main} relies on two major components: (i) text-to-CTR prediction, and (ii) semantically similar ads retrieval.
%\yifan{semantically similar ads retrieval (to be symmetric)}.
Both are challenging problems inherently linked to contextual understanding of ad text. For example, both `Get help from the BEST Home Security System!!!' and `Need help with Social Security Benefits? Call Now!' have the word \textit{security} but the contexts are quite different. The similar ads neighborhood for the two ads, and the average CTRs in their neighborhood can also be different due to the difference in ad categories.
Compared to bag-of-words and word2vec representations, BERT~\cite{bert} based text encoders are particularly helping in understanding such context. Leveraging this observation, for CTR prediction, we fine tune BERT for CTR prediction with ad impressions and clicks data. For similar ads retrieval, we fine tune a sentence-BERT model \cite{sbert_paper} trained for semantic-textual similarity. In particular, we introduce a weak labeling method for ad text pairs using the ad campaign setup information (\emph{e.g.}, two ads from the same advertiser, and the same category label are treated as similar). Such ad-specific fine tuning is effective
in differentiating ads in different categories but with high word overlap.
For example, `Online store for shoes!' and `Online store for pets!' are both online shopping ads and maybe be inferred as similar by a generic textual-similarity-model, but can be differentiated after ad-specific fine tuning based on our weak labeling method.
In TSI, similar ads are not only used to decide whether an input ad is weak or strong, but are also used as suggestions for improvement. It is plausible to use text generation models to suggest the refined version of the input ad text (as in \cite{cikm2020_createbetterads, microsoft_ad_generation_kdd19}). Although there has been significant progress in text generation, current generation methods for ads are not perfect all the time \cite{cikm2020_createbetterads}. They can suffer from hallucinations (inserting irrelevant words), repetitions which affect fluency of the generated text, memorization (\emph{e.g.}, memorizing an ad from an advertiser and generating it verbatim), and high latency which may negatively impact the perception of an advertiser towards the ad platform. Due to such limitations, in this paper we adopt the similar ads approach for refinement suggestions.

Although the primary motivation behind TSI lies in helping small businesses with weaker ads, the core idea 
(as described above) essentially facilitates learning across all advertisers in the ad platform. In the online advertising industry, account managers are common interfaces between large advertisers (clients) and the ad platform. Account managers may launch multiple ad text variations (A/B tests for the same product being advertised), and study their performance over time to infer which text works best. For example, the dating ad insight that mentioning \textit{free sign up} and the \textit{city} (as illustrated in Figure~\ref{fig:pull_figure_main}) can boost the CTR may be inferred by a particular advertiser (and associated account managers). Advertisers may churn out over time due to numerous reasons (\emph{e.g.}, advertising budget cuts), and their learnings can be hard to track manually. In this context, TSI offers the ability to automatically learn across advertisers in a scalable manner (\emph{i.e.}, even for a churned-out advertiser, if the ad text has high predicted CTR, it can be used for relevant TSI suggestions).

Our main contributions can be summarized as follows.
\begin{enumerate}
    \item We introduce a BERT based ad text-to-CTR prediction model and train it using clicks and impressions data from Verizon Media advertisers. The model provides significant AUC lift on top of bag-of-words based models ($1.7\%$ AUC lift in the cold-start setting with new advertisers in the test set). The model also has the ability to ingest publisher information (\emph{i.e.}, the site where the ad will be shown) to provide publisher specific CTR predictions.
    \item We introduce a sentence-BERT based semantic-ad-similarity model to compare two ad texts. The model training leverages ad campaign setup information (advertiser ID and category) to create weak labels for pairs of ads. The model achieves a precision@5 of $0.9$ (and precision@1 $0.93$) in terms of retrieving related ads from the same product category as the input ad. This is $5.8\%$ better than (unsupervised) TF-IDF, word2vec and SBERT based baselines. We use approximate nearest neighbours (ANN) to enable fast retrieval of similar ads from a pool of ~50k ads in less than 0.5 seconds in a production environment.
    
   \item In addition to TSI for English ads from the US, we study TSI for ads from Taiwan and Hong Kong, thereby developing multilingual extensions of the text-to-CTR and semantic-ad-similarity models.%The CTR prediction AUC for the Taiwanese model is comparable to the English one, and the precision@1 is ~0.8 using FastText embeddings (compared to 0.93 for the English model).
   
   \item We tested TSI online as a feature for both onboarding Verizon Media (Yahoo) DSP advertisers (with English ads only), and internal account teams. Our implementation had sub-second end-to-end latency in the DSP UI, providing its users feedback in real time. The account teams were strongly positive about the results (with an average rating of 4.3/5 across $15$ different strategists). Advertisers onboarding via the Verizon Media DSP UI also had positive interactions. In an online test from mid-January to mid-February 2021, $25.12\%$ of DSP advertisers (with English ads) seeing recommendations changed their ads based on ad text seen in TSI suggestions, and such adopters collectively observed a $27\%$ CTR lift compared to non-adopters who were exposed to recommendations. To the best-of-our-knowledge, the TSI feature for providing real time feedback on ad text along with suggestions is the first of its kind among major ad platforms, and it is still available for all Verizon Media advertisers.
\end{enumerate}

The remainder of the paper is organized as follows. Section~\ref{sec:related} covers related work, and Section~\ref{sec:architecture} give an overview of the architecture. We cover our proposed methods for ad text-to-CTR in Sections~\ref{sec:ctr_prediction}. Similar ads retrieval and ad text anonymization are covered in \ref{sec:similar_ads_retrieval}.
Experimental results are covered in Section~\ref{sec:results}, and we end with paper with a discussion in Section~\ref{sec:discussion}.

\section{Related Work} \label{sec:related}

\subsection{Online advertising}
In a typical online advertising setup \cite{mappi_CIKM,Google_FTRL,gemx_kdd}, advertisers design creatives (ad text and image) with the help of creative strategists to target relevant online users
visiting websites (publishers) associated with the ad platform.
The effectiveness of their creatives is measured via metrics such as click-through-rate (CTR $= \frac{clicks}{impressions}$). CTR prediction models \cite{Google_FTRL,mappi_CIKM} are trained to predict the probability of a click given an impression event (and features typically spanning user, publisher, context, device, ad creative, and advertiser).
In this paper, we focus on click prediction models using only the textual information in an ad creative, and the publisher where the ad is being shown. The high level goal is to capture learnings across advertisers to surface best ad text practices, and give contextual creative guidance to advertisers.
In this paper, we do not consider conversions data, and just focus on clicks data, since in most cases, clicks-data (in aggregated form) is accessible by the ad platform for making system-wide improvements.

\subsection{Ad creative image and text understanding}
Understanding ad images and text for the purposes of  
ad creative optimization is an area of active research.
While A/B tests with a large pool of creatives to efficiently learn which creative works best (popularly known as dynamic creative optimization in the industry) \cite{explore_exploit_li, RL_RTB} are a common practice in the industry, recent works \cite{self_recsys2019,www20_joey,cikm2020_createbetterads, cvpr_kovashka} have focused on models to understand ad creatives for gathering insights and automating the ad creation process.
Understanding content in ad images was studied in \cite{cvpr_kovashka,kovashka_eccv2018}, where manual annotations were gathered from crowdsourced workers for: ad category, reasons to buy products advertised in the ad, and expected user responses given the ad.
Leveraging the dataset in \cite{cvpr_kovashka}, \cite{self_recsys2019, www20_joey} studied recommending keyphrases for guiding a brand's creative design using the Wikipedia pages of associated brands.
In \cite{kdd2021_visualtextrank}, semantically similar ads were leveraged to automate ad image search given ad text.
In the context of ad text generation, an encoder-decoder model for generating ad text based on an advertiser's webpage was proposed in \cite{microsoft_ad_generation_kdd19}, and \cite{cikm2020_createbetterads} focused on refining input ad text using generation models. However, current ad text generation models are not perfect \cite{cikm2020_createbetterads}, and suffer from hallucinations, lack of fluency, and high latency. Due to such reasons, we focus on a similar ads approach for refinement suggestions in this paper.
%We formulate ad text generation as a sequence-to-sequence prediction task, which is common in natural language processing problems like machine translation and abstractive summarization. State-of-the-art performance in machine translation is typically obtained with %deep neural networks using
%an encoder-decoder neural architecture with attention \cite{luong2015attention}. In abstractive summarization, where both the source and target sequences are in the same language, an additional mechanism to copy input tokens to the output sequence has proven to be beneficial \cite{see_pointer_generator}. The main differences between our work and \cite{microsoft_ad_generation_kdd19} lie in: (i) studying ad refinement as opposed to generating an ad from scratch, (ii) the use of A/B test data across advertisers to train refinement models.

\subsection{Contextual text embeddings}
Traditionally, many language tasks such as translation are handled using
recurrent neural networks, combined with attention mechanism. This reflects the fact that we tend to read a sentence from left to right. However, human also read words within context of other words, some of them could be quite far apart, instead of only left to right or right to left in a mechanical way. BERT is a recently proposed language model \cite{bert}, and has achieved state-of-the-art performance on a number of natural language understanding tasks. It makes use of the Transformer \cite{transformer} encoder, a self-attention mechanism that learns contextual relations between words (or sub-words) in a text. As opposed to directional models (such as RNN, LSTM and GRU), which read the text sequentially (left-to-right or right-to-left), the Transformer encoder reads the entire sequence of words at once. Therefore it is considered bidirectional, but it is more accurate to say that it is non-directional. This characteristic allows the model to learn the context of a word based on all of its surroundings (left and right of the word). The success of BERT has sparked a subfield (BERTology) which includes RoBERTa \cite{roberta}, ALBERT \cite{albert}, and ERNIE \cite{ernie}, etc. All these methods fall into the category of contextual text embedding networks. In this paper, we leverage pre-trained BERT models for both CTR prediction and retrieving similar ads given an input ad text.
\section{Architecture} \label{sec:architecture}
\begin{figure}[]%!htb
\centering
  \includegraphics[width=1 \columnwidth]{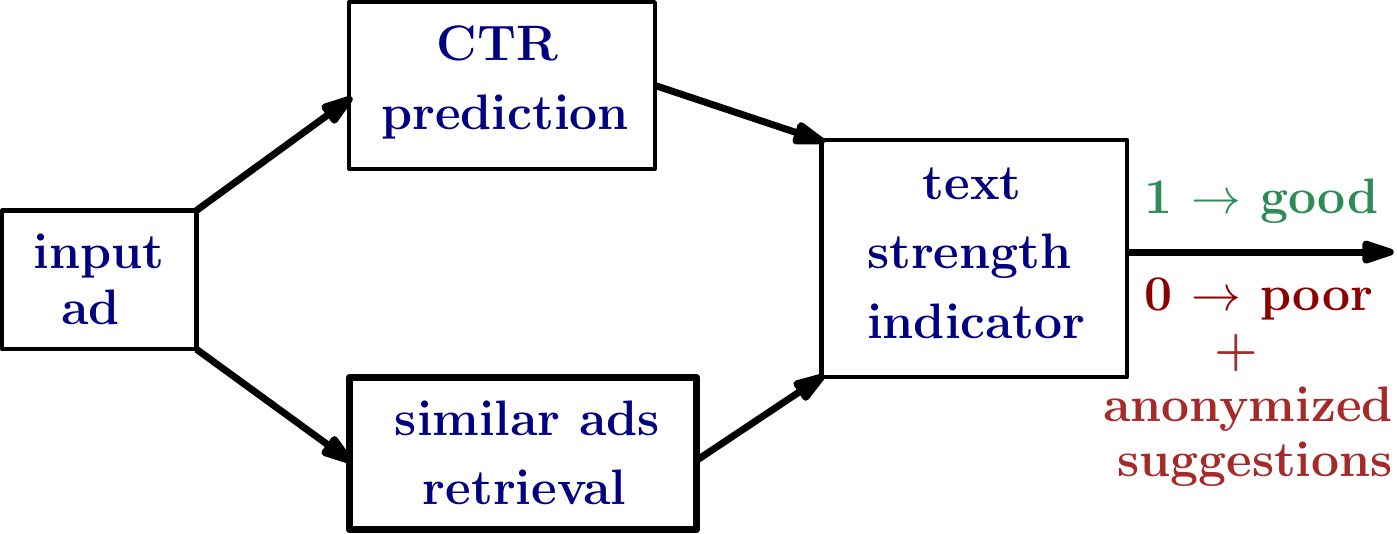}
  \caption{Overall architecture. %\yifan{spell out "anon."}
  }
  \label{fig:architecture}
\end{figure}

The overall architecture for TSI is shown in Figure~\ref{fig:architecture}. As shown, calls to the CTR prediction model and similar ads retrieval model can be made in parallel for the input ad text, and their results can be combined to obtain the final TSI score (with suggestions).
The main components can be briefly described as follows.
\paragraph*{CTR prediction model} We propose a BERT \cite{bert} based CTR prediction model which takes in the ad text (\emph{e.g.}, concatenation of ad title, ad description, and call to action in the case of Verizon Media advertisers), and the publisher (for which CTR is to be estimated).
We take the pre-trained uncased BERT base model released by Google \cite{bert}, and fine-tune it with ad impressions and clicks data.
%For non-English ad text CTR prediction, we consider the corresponding (language specific) BERT model for fine tuning.
Details of the CTR prediction model are in Section~\ref{sec:ctr_prediction}.

\paragraph*{Similar ads retrieval model}
For similar ads retrieval, we assume a pool of existing ads ($\sim$ $50k$ for our
%\yifan{(do we need to disclose our production setting? or just change online to ``our"?)}
experiments).
Each ad in the pool has its original ad text, its anonymized version (without references to a brand), and the associated pCTR (predicted click-through rate, computed in advance from the CTR prediction model). The goal of similar ads retrieval is to find top $k$ relevant (semantically similar) ads from the pool, given an input ad text. In this paper, we propose a semantic-ad-similarity model which produces ad text embeddings, and compares two ads using cosine similarity between their ad text embeddings. Our semantic-ad-similarity model is based on sentence-BERT \cite{sbert_paper} fine tuned with weak similarity labels from the ads data (additional details in Section~\ref{sec:similar_ads_retrieval}).

\paragraph*{TSI} The TSI block needs the results for CTR prediction and similar ads retrieval for a given input ad.
Intuitively, the TSI block checks whether the input ad's pCTR is low compared to pCTRs of other ads in its semantic neighborhood. 
If there are a significant number of superior ads (above a predefined pCTR difference threshold) in its neighborhood, the input ad is labeled \textit{weak} ($0$), else it is labeled strong ($1$). In addition, for a weak input ad, anonymized versions of superior ads are shown as improvement suggestions. Additional details on the TSI logic are in Section~\ref{sec:tsi}.
\section{Text-to-CTR prediction} \label{sec:ctr_prediction}
%We explored several models for CTR prediction, including linear models like logistic regression (LR) and naive Bayes logistic regression (NBLR) \cite{nblr}, and a deep learning model like BERT \cite{bert}.
The objective for our text-to-CTR prediction model was to come up with a click probability as defined below.
\begin{align}
pCTR = \mathbb{P}\left( click | ad\;text, \; publisher \right),
\end{align}
where $pCTR$ is the model's predicted click-through rate, $click\in\{0,1\}$ is a binary label indicating click ($1$) or no click ($0$) for an ad impression event, $ad\;text$ is the concatenation of ad title, description and call to action, and $publisher$ indicates the publisher where the ad is being shown (\emph{e.g.}, Yahoo Mail). The loss function for our CTR prediction model is weighted binary cross entropy as defined below.
\begin{align} \label{eq:CTR_loss}
\ell = -\frac{1}{\sum_{i=1}^N w_i}\sum_{i=1}^N w_i \left[y_i \log (p(y_i))+(1-y_i) \log (1-p(y_i))\right],
\end{align}
where $w_i$ is the weight for each sample (explained below). Each ad is treated as two samples, one with label $1$ meaning it is clicked, and another with label $0$ meaning it is not clicked. Therefore $N$ in the loss function \eqref{eq:CTR_loss} will be twice the number of our ads. Depending on the label, the weight assigned to each sample can be either the number of clicks if the label is $1$, or the number of no clicks (impressions-clicks) if the label is $0$.
%% SHAUNAK: ADD note on confounding factors later?
% \begin{figure}[]%!htb
% \centering
%   \includegraphics[width=1 \columnwidth]{./images/BERT_arc.png}
%   \caption{BERT model. \yifan{this seems like the pretraining part. Should we illustrate the fine-tuning part?}}
%   \label{fig:bert}\changwei{I am not sure about this considering BERT is so well-known now. I actually feel it might be fine to remove the pre-training architecture.}\yifan{Agree, that's why I thought we should only include the fine tuning part}
% \end{figure}

\begin{figure}[]%!htb
\centering
  \includegraphics[width=0.7 \columnwidth]{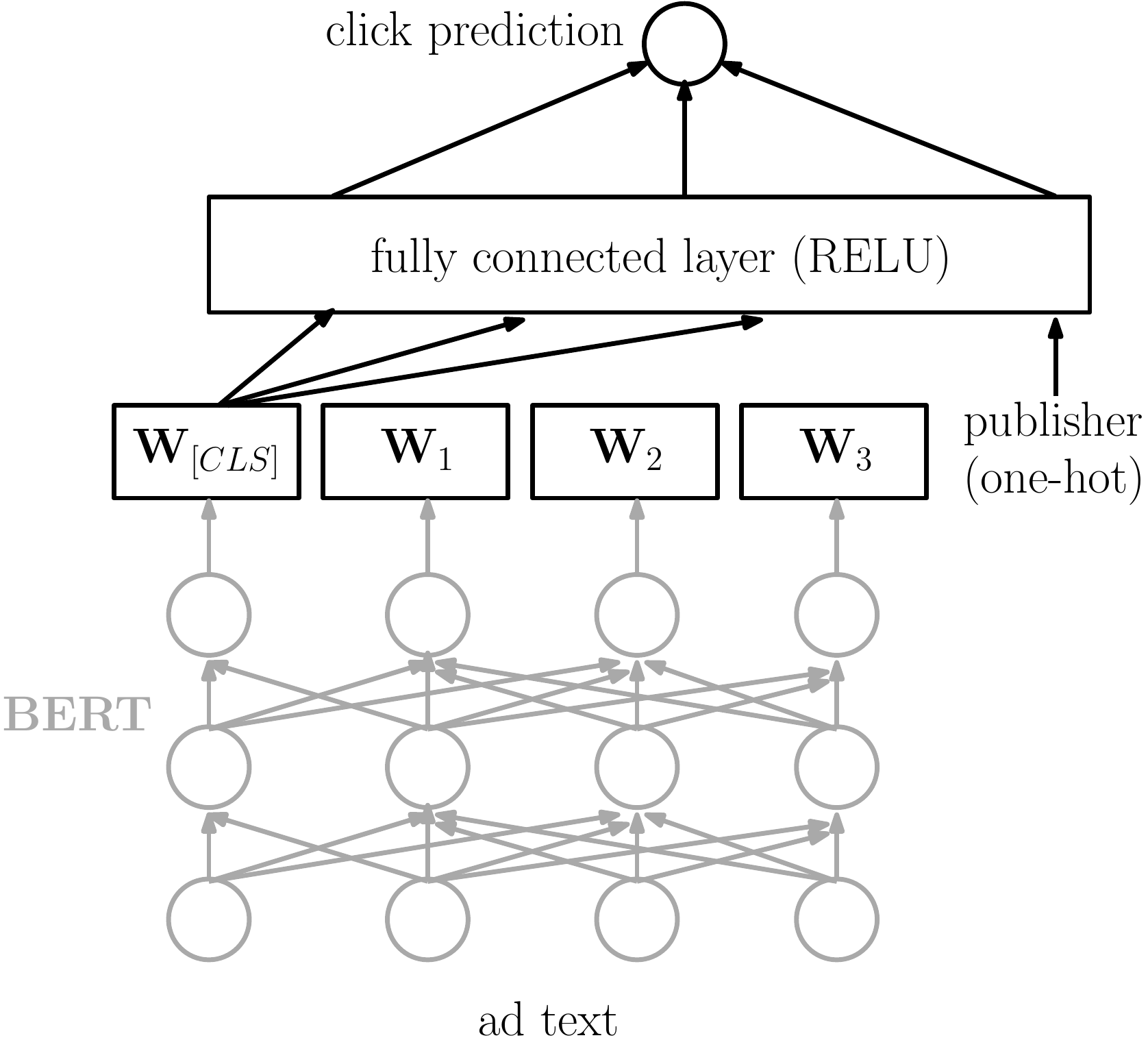}
  \caption{BERT fine-tuning for CTR prediction.}
  \label{fig:bert_ft}
\end{figure}
The architecture for our proposed BERT fine-tuning (for CTR prediction) is shown in Figure \ref{fig:bert_ft}. The BERT model is essentially the encoder part of a Transformer network \cite{transformer}. BERT is a self-supervised language model which learns contextual relations between word pieces by training on multiple tasks including masked token prediction and next segment prediction.
For fine-tuning BERT in our CTR prediction setup, we use the output from \emph{pooled\_output}\footnote{https://github.com/google-research/bert} layer as the embedding for the text. The \emph{pooled\_output} layer takes the first token ([CLS] token) as the input followed by a dense layer with \emph{Tanh} activation. To improve the CTR prediction, we further leverage the publisher feature by concatenating the text embedding with the one-hot encoding of publishers (restricted to $max\_pub = 13$ most popular publishers), and then stack it with a dense layer of 64 nodes.
In our experiments (details in Section~\ref{sec:ctr_prediction_results}), we observed that fine-tuning of the BERT model leads to a significant improvement in CTR prediction AUC compared to the case without any fine-tuning; for additional training details see the Appendix (reproducibility notes).
\section{Similar ads retrieval} \label{sec:similar_ads_retrieval}
 In this section, we first go over our similar ads retrieval model
 based on semantic-ad-similarity in Section~\ref{sec:supervised_sbert}, followed by Section~\ref{sec:anon} where we explain our method for anonymizing ads (which may be shown as TSI suggestions to advertisers after similar ads retrieval).  

\subsection{Semantic-ad-similarity}\label{sec:supervised_sbert}
For similar ads retrieval, we assume a set (pool) $\mathbf{P}$ of existing ads at least in the order of tens of thousands ($ |\mathbf{P}| \sim 50k $ ads in our online experiments), and we need to retrieve top $k$ relevant ads from $\mathbf{P}$ given an input ad. The relevance of an ad $\mathbf{a} \in \mathbf{P}$ with respect to the input ad $\mathbf{a}_{in}$ is defined as:
\begin{align} \label{eq:cosine_similarity}
    relevance(\mathbf{a}_{in}, \mathbf{a}) = < \phi(\mathbf{a}_{in}), \phi(\mathbf{a})  >,
\end{align}
where $\phi(\mathbf{a})$ is the (normalized) embedding of $\mathbf{a}$ based on its ad text, and $< \phi(\mathbf{a}_{in}), \phi(\mathbf{a}) >$ denotes the cosine similarity between the embeddings of $\mathbf{a}_{in}$ and $\mathbf{a}$. We use such cosine-similarity based relevance ranking framework (as opposed to having a neural network, \emph{e.g.}, DRMM \cite{drmm_topk} score a pair of ad texts) due to the scale of our setup; it is convenient to use approximate nearest neighbor approaches \cite{spotify_ann} with cosine similarity as well. This choice is echoed by recent works on sentence-BERT based semantic-textual-similarity models \cite{sbert_paper}. With \eqref{eq:cosine_similarity} in place, the problem of similar ads retrieval boils down to learning embeddings $\phi(\cdot)$ which focus on the context of the ad (\emph{i.e.}, product being advertised).
To obtain such embeddings in a supervised manner leveraging ads data, we propose: (i) a weak labeling method for pairs of ad text (details on Section~\ref{sec:weak_labels}), and (ii) fine-tuning a sentence-BERT model with such weakly labeled pairs (details in Section~\ref{sec:sbert_finetuning}).
\begin{figure}[]%!htb
\centering
  \includegraphics[width=0.8 \columnwidth]{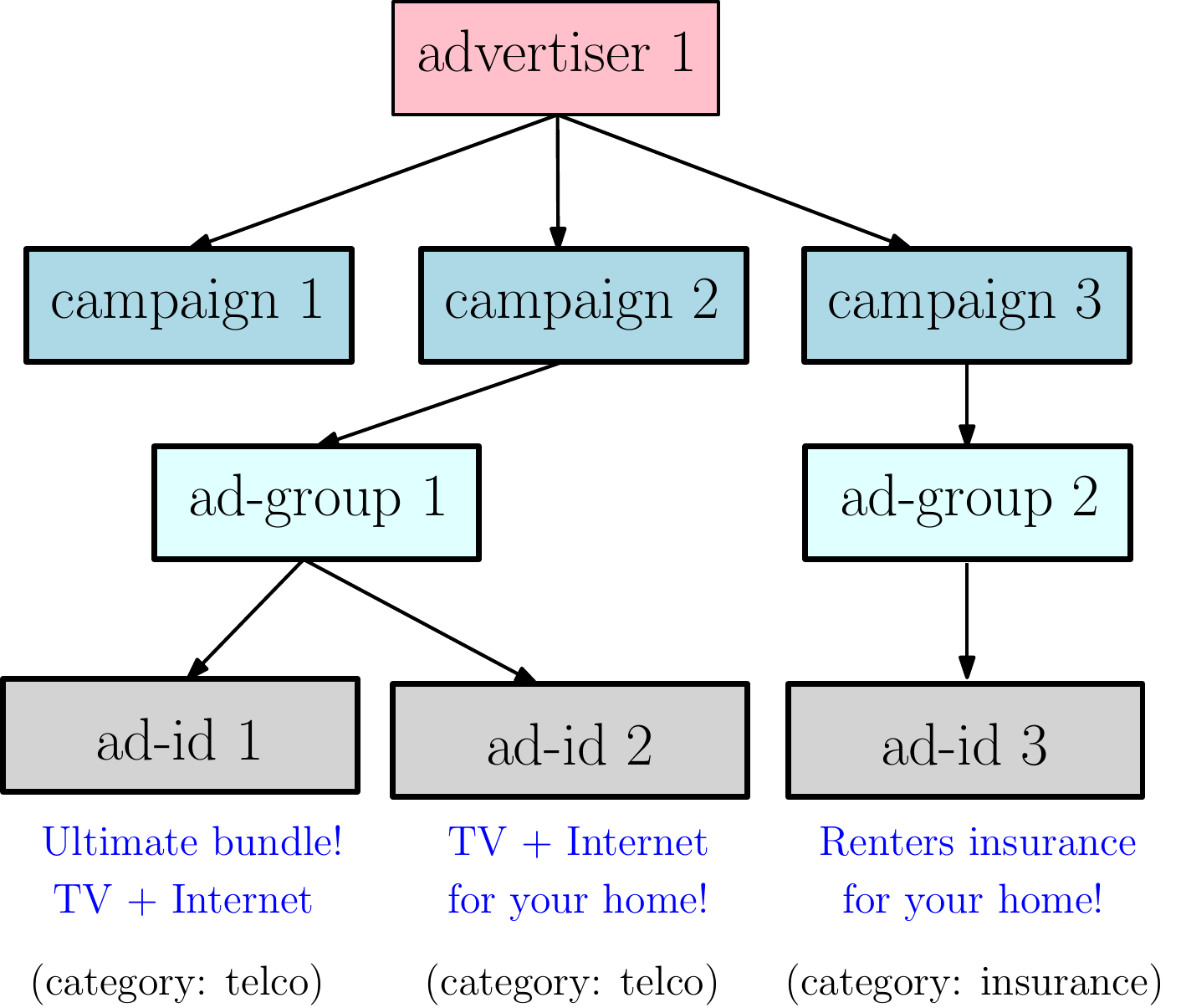}
  \caption{Illustrative example of ad campaign setup in Verizon Media. An advertiser can have multiple campaigns, leading to multiple ad IDs (tied to ad text). The user targeting criteria is usually set at an ad group level, and the advertiser can simultaneously advertise in multiple product categories. In the example, ad IDs 1 and 2 can be treated as similar.}
  \label{fig:cmp}
\end{figure}

\subsubsection{Weak labeling using ads data} \label{sec:weak_labels}
%As such, \yifan{As what? delete "As such"?}
There is no prior data set with pairs of ads labeled as similar. However, it is plausible to exploit the structure of campaign setup in ad platforms to come up with \textit{weak} similarity labels for ad text pairs. Figure~\ref{fig:cmp} shows an example of how an ad campaign may be setup: the advertiser can have multiple campaigns (typically tied to product category but not necessarily). Each campaign can have ad groups associated with it where the user targeting criteria (\emph{e.g.}, location targeting) is specified, and finally within an ad group there may be multiple ad IDs (each with different text or image but typically tied to the same product). The category (\emph{e.g.}, IAB category \cite{iab_com}) associated with each ad ID may be self-declared by the advertiser (noisy) or may be inferred via a category classifier\footnote{In our experiments, we consider $\sim 20$ categories corresponding to the top-level IAB categories \cite{iab_com}. This is not fine grained enough to identify ads advertising the same product type, \emph{e.g.}, both shoes and shirts ads may be in the same (apparel) category.}. Following the intuition in Figure~\ref{fig:cmp}, we weakly label a pair of ads $(\mathbf{a}_i, \mathbf{a}_j)$ from ads pool $\mathbf{P}$ as positive (similar) if both of them are from the same advertiser, and have the same category. If they have different categories, and are from different advertisers then the pair is marked negative (dissimilar). We arrived at this combination (\emph{i.e.}, choosing the advertiser to be same as opposed to the same campaign or ad group) via offline experiments (discussed in Section~\ref{sec:results}). Given a large advertiser base, our weak labeling data can easily generate sufficient data to fine-tune a sentence-BERT model as discussed below.

\subsubsection{Sentence-BERT fine-tuning with weak labels} \label{sec:sbert_finetuning} 
\begin{figure}[]%!htb
\centering
  \includegraphics[width=0.5 \columnwidth]{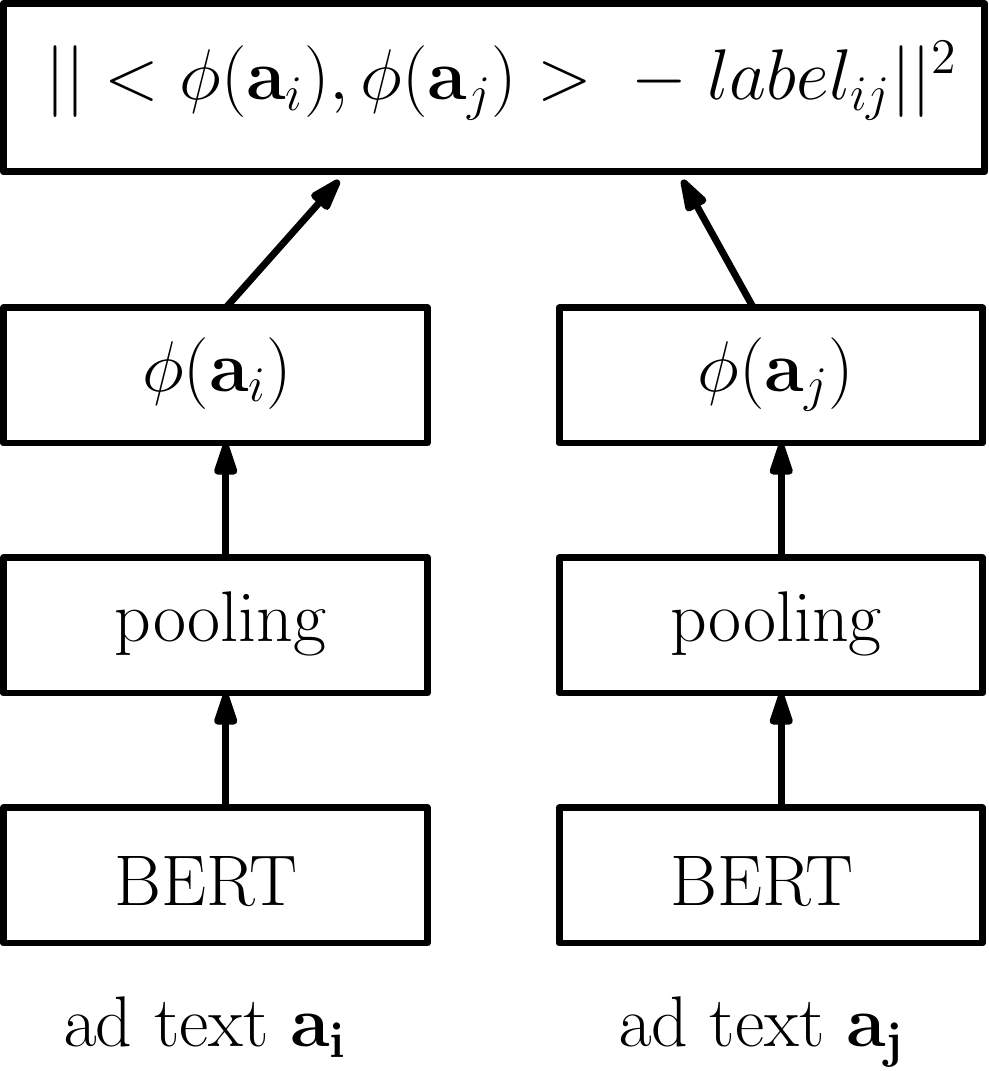}
  \caption{Training semantic-ad-similarity model (SAS-SBERT) by fine-tuning sentence-BERT with cosine similarity loss on weakly labeled ad pairs.}
  \label{fig:sbert}
\end{figure}

Given weakly labeled pairs of ad text, we build on the work in \cite{sbert_paper} to obtain fine tuned sentence-BERT representation for an ad text.
Sentence-BERT adds a pooling operation to the output of BERT to derive a sentence embedding as shown in Figure~\ref{fig:sbert}.
We take a pretrained sentence-BERT model for semantic-textual-similarity (with mean-pooling), and fine-tune it with the weakly labeled ad text pairs. For fine-tuning, we use
use mean-square (regression) loss for a pair $( \mathbf{a_i}, \mathbf{a_j} )$ as defined below:
\begin{align} \label{eq:sbert_cosine_loss}
 loss_{ij} = ||< \phi(\mathbf{a}_{i}), \phi(\mathbf{a}_j) > - label_{ij}  ||^2,
\end{align}
where weak $label_{ij} \in \{1,-1\}$.
As we describe later in Section~\ref{sec:results}, this loss function yielded the best (offline) results compared to cross-entropy loss, and triplet loss (by using triplets of positive, positive, and negative text as in \cite{sbert_paper}). We will refer to our proposed model (as desribed above) as SAS-SBERT (Semantic-Ad-Similarity SBERT).
%\yifan{We call this algorithm SAS-SBERT (Semantic-Ad-Similarity SBERT)}
\begin{comment}
\paragraph*{SBERT}
change: SBERT adds a pooling operation to the output
of BERT / RoBERTa to derive a fixed sized sentence embedding. We experiment with three pooling strategies: Using the output of the CLS-token,
computing the mean of all output vectors (MEANstrategy), and computing a max-over-time of the
output vectors (MAX-strategy). The default configuration is MEAN. In order to fine-tune BERT / RoBERTa, we create siamese and triplet networks (Schroff et al.,
2015) to update the weights such that the produced
sentence embeddings are semantically meaningful
and can be compared with cosine-similarity.Triplet Objective Function.
Given an anchor
sentence a, a positive sentence p, and a negative
sentence n, triplet loss tunes the network such that
the distance between a and p is smaller than the
distance between a and n. Mathematically, we
minimize the following loss function:
max(||sa − sp|| − ||sa − sn|| + , 0)
with sx the sentence embedding for a/n/p, || · ||
a distance metric and margin . Margin  ensures
that sp is at least  closer to sa than sn. As metric
we use Euclidean distance and we set  = 1 in our
experiments.
\end{comment}

\subsection{Anonymizing ad text} \label{sec:anon}
In our setup, retrieved similar ads may be shown as suggestions for improvement for an input ad text. To discourage copying ads verbatim, and to prevent users of TSI from identifying a particular brand (which may lead to biases in TSI adoption), we anonymize the ads pool $\mathbf{P}$ by removing references to a brand in ad text. For anonymizing ad text, we use a block list based approach (with brand and product names) coupled with off-the-shelf named entity recognition (NER) tools \cite{spacy}.
%For updating the block list of brands to be anonymized, we use a combination of off-the-shelf named entity recognition (NER) tools and periodic human review.
\section{TSI} \label{sec:tsi}
\begin{figure}[]%!htb
\centering
  \includegraphics[width=0.9 \columnwidth]{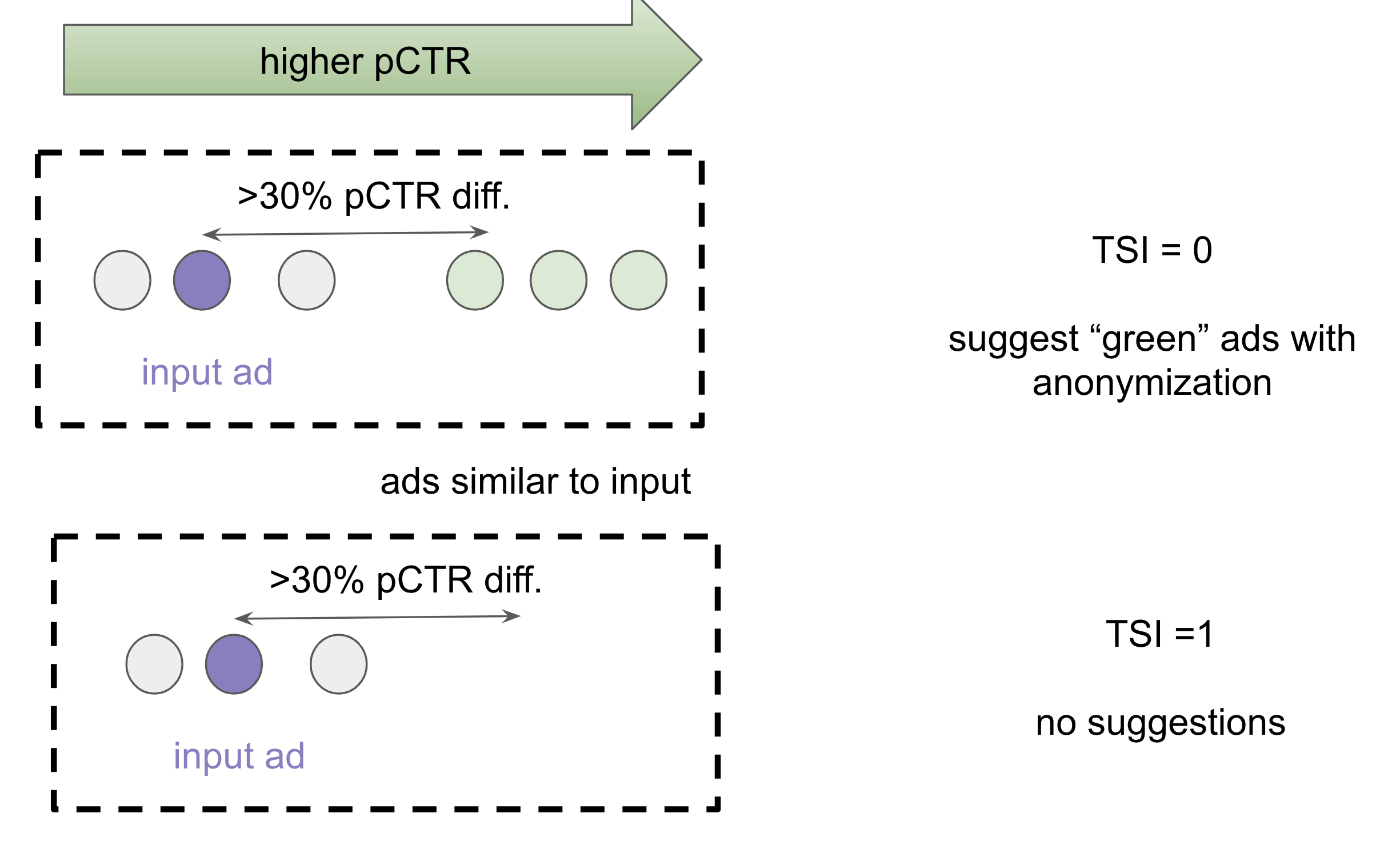}
  \caption{Illustrative example for TSI. TSI is 0 when there are sufficient ads above the input ad (relative pCTR difference above threshold $\Delta = 30\%$).}
  \label{fig:tsi_idea}
\end{figure}

TSI's core motivation lies in detecting poorly written ads, and providing suggestions for improvement.
However, instead of rules (thresholds) defined on absolute pCTR values, we focus on the (semantic) neighborhood of the input ad text (where its neighbors are the top-$k$ retrieved similar ads from the pool $\mathbf{P}$).
The goal of TSI is to provide an estimate of the goodness (pCTR) of the input ad relative to its neighbours (similar ads).
For our experiments, we used the following rule based algorithm for TSI.
\begin{enumerate}
\item Retrieve top $k=5$ neighbors for the input ad, and obtain the pCTRs of the input and the $k$ neighbors,
\item compute median pCTR of ads above input ad in pCTR, and
\item if the above median
%\yifann{> $\Delta \% = 30\%$}
is more than $\Delta\%$ (we use $\Delta\%= 30\%$) of the input ad pCTR, TSI = 0 (means input is weak), else TSI = 1 (input is strong). The corresponding neighbors (above $\Delta \% $ in pCTR compared to input) are shown as ad text suggestions (anonymized versions) for improving the input ad.
\end{enumerate}
Figure~\ref{fig:tsi_idea} illustrates the neighborhood-based TSI as described above.
As we describe later in Section~\ref{sec:tsi_offline_reco_rate}, the choice of $\Delta$ is crucial in determining the chances of an ad text getting TSI=0 score (higher $\Delta$ can lead to fewer neighbors with pCTR better than the threshold, and a lower chance of getting TSI=0 score).
%Also, instead of computing the median pCTR above, we could have used the mean pCTR; we preferred median over mean as the latter was a stricter criteria.

\section{Results} \label{sec:results}
%In this section we first 
\subsection{Datasets} \label{sec:datasets}
For offline experiments with our CTR prediction model and semantic-ad-similarity model, we collected data for one month from Verizon Media ad platform. For each ad-ID, the following fields were collected: ad title, ad description, call to action (CTA) text, clicks, impressions, and publisher. The ad text was defined as the concatenation of ad title, description and call-to-action with a period mark.
The publishers were limited to top $15$ publishers by impression count.
Furthermore, we sampled the data collected (from US advertisers) to prepare a set of $50,000$ ad IDs spanning over $4,000$ advertisers.
For the CTR prediction model, two datasets were created: (i) warm-start, (ii) cold-start. Warm start dataset has train-test-validation random splits with the possibility of ads from the same advertiser occurring in multiple splits. In the cold start dataset, an advertiser can occur only in one split (\emph{i.e.}, the test advertisers do not occur in train data).
For experiments with ads from Taiwan and Hong Kong (Chinese ad text), we repeated the above procedure to get a sample of $15,000$ ads for Taiwan, and $5,000$ ads for Hong Kong. 
%The data consist of 298,229 ads from 3,696 advertisers, and 102 unique CTAs.

\subsection{CTR prediction}\label{sec:ctr_prediction_results}
We evaluate our fine-tuned BERT model (as in Section~\ref{sec:ctr_prediction}) with two experimental settings: warm-start (random split) setting and cold-start setting. For random split setting, we randomly do a 80/6/14 train/validation/test split. For cold-start setting, we group ads based on their advertisers, and split into train/validation/test.
Before going over the results, we explain below our metrics and baselines.
%by taking ads from 2,956/222/518 advertisers respectively.
% BATCH_SIZE = 32
% LEARNING_RATE = 2e-5
% NUM_TRAIN_EPOCHS = 20
%%% ADD 
\paragraph*{Evaluation metrics}
Our CTR prediction model can be used not just for predicting the CTR of an ad, but also as an ad text ranker for advertisers intending to rank multiple ad text variations by pCTR.
Therefore, we use AUC to measure how our estimated CTRs perform in terms of classifying clicks and not-clicks, and use Kendall Tau-b coefficient (KTC) \cite{ktc} and Spearman's rank correlation coefficient (SRCC) \cite{srcc} to measure how well the estimated CTR's order aligns with ad text ranking derived from the ground truth CTRs.

KTC is a measure of the correspondence between two sets of rankings $x$ and $y$. It is defined as
\begin{equation}
    \tau_B = \frac{2(P-Q)}{\sqrt{[n(n-1)-T][n(n-1)-U]}}
\end{equation}
where $P$ is the number of concordant pairs, $Q$ the number of discordant pairs, $T$ the number of pairs with a tie only in ranking array $x$, and $U$ the number of ties only in $y$, and $n$ is the number of entries in $x$ and $y$. If a tie occurs for the same pair in both $x$ and $y$. Values of KTC close to 1 indicate strong agreement, values close to -1 indicate strong disagreement. SRCC is a nonparametric measure of the monotonicity of the relationship between two datasets $x$ and $y$. Like KTC, its value also varies between -1 and 1 with 0 implying no correlation. Correlations of -1 or 1 imply an exact monotonic relationship. Positive correlations imply that as $x$ increases, so does $y$. Negative correlations imply that as $x$ increases, $y$ decreases.

Note that in calculating AUC, we treat each impression as a sample separately. Two impressions of the same ad could be clicked and not-clicked respectively, but our model could only give one probability for both. Hence the AUC has an upper-bound, which is obtained by setting the pCTR to be the ground truth CTR, that is significantly less than 1. Considering this, we introduce an additional metric named relative AUC to evaluate how close our estimated CTR is to the actual CTR. The relative AUC is defined as
\begin{equation}
    \text{Relative AUC} = \frac{\text{AUC}}{\text{Upper-bound AUC}}
\end{equation}
In measuring KTC and SRCC, on the other hand, we
treat each ad as one sample, and measure how well our predicted CTRs are, in terms of giving the same ordering as the actual CTRs.
%In addition, we believe that the publisher of an ad can have a direct impact on the CTR (Our UI actually provides publisher as an optional input). We therefore evaluate the usefulness of publisher information in our experiments.
\paragraph*{Baselines} As baselines, we considered logistic regression (LR), naive Bayes logistic regression (NBLR) \cite{nbsvm}, and BERT without fine-tuning. 
Both LR and NBLR are based on bag-of-words text representation; NBLR is a model similar to Naive Bayes SVM \cite{nbsvm}, with the SVM part replaced by logistic regression.

\paragraph*{Offline results} Table \ref{tab:ctr_pred} shows offline results for the warm start setting (random split).
%Our experimental results on data of random split are shown in .
As can be observed, NBLR performs better than logistic regression (LR) due to the introduction of the naive Bayes ratio. Fine-tuned BERT\footnote{For BERT fine-tuning, we initialize our model using the uncased BERT base checkpoint. The batch size is set to 32, the learning rate is set to $2\times 10^{-5}$, and the seq\_length is set to 100. We run 20 epochs to fine-tune the parameters.} and NBLR both perform well. While BERT is better in terms of AUC, NBLR achieves better KTC and SRCC. By comparing row 3 and row 4, we find that BERT model without fine-tuning performs much worse than fine-tuned BERT. It is therefore necessary to fine-tune BERT for the CTR prediction task. The results in row 3 and row 5 indicate that CTR is biased by the publisher\footnote{In addition to the publisher, there can be other confounding factors affecting the CTR, like the choice of ad image, location of user, and user's device. We focus on publisher in this paper as it is a possible input in our end application, but it is straightforward to integrate the other features listed above in our framework.} in the case of BERT fine-tuning. We observed the same publisher bias when a linear NBLR model was trained using set 2 features, and the performance was similar to that of the fine-tuned BERT model. Based on this observation, we include publisher as an additional input (can be optional) in our online tests with the TSI feature. We also explore the ensemble of NBLR with BERT. However, the results in row 6 of Table \ref{tab:ctr_pred} indicate that an ensemble model does not provide any performance gains. In addition, we want to stress that although the best AUC we get here is only 0.735, it is worth mentioning that the upper bound AUC for the test data (by using the ground truth CTR) is 0.7812.
\begin{table}[h]
\setlength{\tabcolsep}{3pt}
    \centering
    \begin{tabular}{l|l|c|c|c|c|c}
    \hline 
Row & Methods & Features	&	AUC	& \vtop{\hbox{\strut Relative}\hbox{\strut AUC}} &	KTC	&	SRCC\\\hline
1 & LR & Set 1 & 0.7240 & 92.67\% &	0.3436 & 0.4872	\\\hline
2 & NBLR & Set 1 & 0.7252 &92.83\% &	0.3891 & 0.5431	\\\hline
3 & BERT & Set 1 & 0.7254 &	92.85\% & 0.3652 & 0.5138	\\\hline
4 & \vtop{\hbox{\strut BERT (no}\hbox{\strut fine-tuning)}} & Set 1 & 0.6862 & 87.83\%& 0.2426 & 0.3550\\\hline
5 & BERT & Set 2 & 0.7351 & 94.09\%&	\textbf{0.4418} & \textbf{0.6118}	\\\hline
6 & BERT+NBLR & Set 2, 1 & \textbf{0.7352} & 94.11\% & 0.4414 & 0.6112\\\hline
     \end{tabular}
    \caption{CTR prediction. Feature set 1 includes ad title, description and call to action features, and set 2 includes ad title, description, call to action, and additional publisher feature. All BERT models are fine-tuned except row 4.}
    \label{tab:ctr_pred}
\end{table}

The results for the cold-start setting are listed in Table \ref{tab:ctr_cold}. The upper-bound AUC obtained by the ground truth CTR is 0.7252. As can be observed, using BERT leads to $1.7\%$ increase in AUC, $0.06$ and $0.08$ increase in KTC and SRCC respectively. We believe this improvement in performance is due to the fact that the BERT model is pre-trained using a large amount of text data, and its language modeling tasks allow BERT to learn lexical similarity in word piece representation. Due to this, BERT is less sensitive to cold-start problem compared with NBLR.
Because of the superior performance of the fine-tuned BERT model in the cold start setting (which is the more likely case for an inexperienced onboarding advertiser), we use the same for our online tests with the TSI feature.
\begin{table}[h]
    \centering
    \begin{tabular}{l|c|c|c|c}
    \hline 
Methods	& AUC &Relative AUC	&	KTC	&	SRCC\\\hline
NBLR & 0.6314 & 87.06\%&	0.2052 & 0.2976	\\\hline
BERT & {\bf 0.6483} & 89.39\%& {\bf 0.2616} & {\bf 0.3758}	\\\hline
     \end{tabular}
    \caption{CTR prediction results in the cold-start setting.}
    \label{tab:ctr_cold}
\end{table}

\paragraph*{Multilingual extension for CTR prediction model}
We extended our CTR predictor to ads from Taiwan and Hong Kong, both of which use traditional Chinese text. The performance of our CTR prediction model is shown in Table \ref{tab:ctr_pred_chinese}. For Chinese ads, we fine-tuned a BERT model pre-trained for Chinese text on Wikipedia (provided by TensorFlow Hub\footnote{https://tfhub.dev/tensorflow/bert\_zh\_L-12\_H-768\_A-12/1}). Our model was fine-tuned using both Hong Kong and Taiwan ads data.
\begin{table}[h]
    \centering
    \begin{tabular}{l|c|c|c|c}
    \hline 
Testing Data	&	AUC	& Relative AUC	& KTC	&	SRCC\\\hline
HK+TW & 0.7597 & 93.66\%	 & 0.4741 &	0.6428\\\hline
HK & 0.6971 & 90.20\% & 0.4302 & 0.6029	\\\hline
TW & 0.7657 & 93.96\%	& 0.4816 &	0.6496\\\hline
     \end{tabular}
    \caption{CTR prediction for Taiwan (TW) and Hong Kong (HK) based ads (Chinese ad text) using fine-tuned BERT.}
    \label{tab:ctr_pred_chinese}
\end{table}
We evaluated performance on three different testing datasets: (i) Hong Kong and Taiwan mixed (HK+TW), (ii) Hong Kong only (HK), and (iii) Taiwan only (TW) data. 
We obtained CTR prediction (using fine-tuned BERT) AUCs comparable to the English ads model ($\sim$0.74) for the Chinese dataset.
%while the AUC for Hong Kong data is lower. 
%This is because Hong Kong data are much smaller in size than Taiwan data, and have a smaller AUC upper-bound.
%The precision@1 for Taiwan data dropped to 0.8 (compared to 0.93 for English ads). We also see a drop of precision@1 compared with Taiwanese ad results.\textcolor{blue}{(Shaunak, do you want to add a brief explanation for the drop in precision@1)}

\subsection{Similar ads retrieval} \label{sec:similar_ads_retrieval_results}
For the purposes of evaluating the similar ads retrieval model (offline), we consider different notions of similarity. With reference to the ad campaign setup hierarchy in Figure~\ref{fig:cmp}, two ads in the same ad group form our strongest notion of similarity, and two ads in the same category form our weakest notion of similarity (same campaign, and same advertiser cases are of intermediate nature). For a given similarity notion (type), we measure the precision@k of the list of $k$ retrieved ads given an input ad.

As baselines, we consider sentence (ad text) representations using TF-IDF, word2vec \cite{w2v} based sentence representation (\emph{i.e.}, mean of word2vec vectors of tokens) as offered in Spacy \cite{spacy}, and the sentence-BERT model \cite{sbert_paper}. Table~\ref{tab:supervised_sbert_vs_baselines} shows the category precision@$k$ (\emph{i.e.}, the fraction of top $k$ similar ads which are in the same category as the input ad) for the baselines versus the proposed model (semantic-ad-similarity using SBERT fine tuned with ads data). 
\begin{table}[h]
    \centering
    \begin{tabular}{|l|c|c|c|c|c|c|}
    \hline 
 similarity	&	cat.	&	cat.	&	cat.	&	cat.	&	cat.	\\
 models	   &	P@1	&	P@2	&	P@3	&	P@5	&	P@10	\\
\hline	
TF-IDF	&	0.89	&	0.85	&	0.83	&	0.79	&	0.74	\\
word2vec	&	0.89	&	0.86	&	0.84	&	0.81	&	0.76	\\
SBERT	&	0.91	&	0.89	&	0.87	&	0.85	&	0.83	\\
\textbf{SAS-SBERT}	&	\textbf{0.93}	&	\textbf{0.92}	&	\textbf{0.91}	&	\textbf{0.9}	&	\textbf{0.89}	\\
     \hline
     \end{tabular}
    \caption{Comparison of category precision@$k$, for different baselines and the proposed semantic-ad-similarity model using fine tuned sentence-BERT (SAS-SBERT).}
    \label{tab:supervised_sbert_vs_baselines}
\end{table}
The results for the proposed model are based on the cosine similarity loss defined in \eqref{eq:sbert_cosine_loss} considering two ads in the same category and from the same advertiser as a weakly positive label while training\footnote{For training SAS-SBERT for English ads, we used the distilbert-base-nli-stsb-mean-tokens model as our initial model \cite{sbert_paper}. In comparison, BERT (base and large), and RoBERTa (base and large) had marginal benefits with longer training time.
The SBERT hyperparameters were: positive-to-negative sampling ratio = 1:30 , batch size = 30, and epochs = 10.
%Other hyperparameters were copied from \cite{sbert_paper} for the STS task.
}.
As shown in Table~\ref{tab:supervised_sbert_vs_baselines}, the proposed model clearly outperforms the baselines (even sentence-BERT); the differences are particularly significant as $k$ increases (the proposed model has $7.2\%$ lift over baselines at $k=10$).
To understand how the proposed model fares for different notions of similarity (following the ad campaign setup in Figure~\ref{fig:cmp}), Table~\ref{tab:supervised_sbert_retrieval} reports the precision at $k$ results (for $k \in \{1,2,3,5,10 \}$) for multiple notions of similarity (\emph{i.e.}, ad group, campaign, advertiser, and category).
%%%%%%%%%%%%%%%%%%%
\begin{comment}
\begin{table}[h]
    \centering
    \begin{tabular}{|l|c|c|c|c|c|c|}
    \hline 
similarity	&	P@1	&	P@2	&	P@3	&	P@5	&	P@10	\\
type	&		&		&		&		&		\\
\hline
ad group	&	0.17	&	0.13	&	0.11	&	0.09	&	0.06	\\
campaign	&	0.2	&	0.16	&	0.14	&	0.11	&	0.08	\\
advertiser	&	0.72	&	0.69	&	0.68	&	0.65	&	0.61	\\
category	&	0.93	&	0.92	&	0.91	&	0.9	&	0.89	\\
     \hline
     \end{tabular}
    \caption{Similar ads retrieval precision@$k$ for the proposed semantic-ad-similarity model (sentence-BERT fine tuned with weak labels) for different notions of similarity.}
    \label{tab:supervised_sbert_retrieval}
\end{table}
\end{comment}
%%%%%%%%%%%%%%%%%%%
\begin{table}[h]
    \centering
    \begin{tabular}{|l|c|c|c|c|}
    \hline 
similarity	&	advertiser-cat	&	campaign-cat	   &	adgroup-cat	\\
type	    &	      labeling	&	labeling           & labeling       	\\
     	    &	          P@1	&		P@1            &	P@1       	\\
\hline
ad group	&	0.17	&	\textbf{0.21}	&	0.2		\\
campaign	&	0.2		&	\textbf{0.25}	&	0.23		\\
advertiser	&	\textbf{0.72}	&	0.69	&	0.68		\\
category	&	\textbf{0.93}	&	0.87	&	0.86	  \\
     \hline
     \end{tabular}
    \caption{Similar ads retrieval precision@$1$ for different notions of similarity, and different weak labeling strategies (\emph{e.g.}, campaign-cat denotes the strategy where two ads with same campaign ID and category are labeled positive).}
    \label{tab:supervised_sbert_retrieval}
\end{table}
Table~\ref{tab:supervised_sbert_retrieval} also shows the impact of different weak labeling strategies on the precison@k for different notions of similarity: for advertiser-cat, two ads with the same advertiser ID and category are labeled positive, while for campaign-cat, and adgroup-cat, the restriction is changed to campaign ID and adgroup ID respectively (instead of advertiser ID). The advertiser-cat combination works best for similarity at an advertiser and category level, while campaign-cat works best for the campaign and ad-group level notion of similarity.
Table~\ref{tab:supervised_sbert_retrieval} also shows that the proposed model retrieves
ads across advertisers during similar ads retrieval. For example, if the category $P@1$ is 0.93, and the advertiser $P@1$ is $0.72$, roughly $0.21$ fraction of test cases have their closest (most similar ad) fetched from a different advertiser (sharing the same category). In addition to the above, we explored cross-entropy loss and triplet loss \cite{sbert_paper} with our best weak labeling strategy (\emph{i.e.}, using advertiser ID and category), and found that cosine similarity loss (as defined in \eqref{eq:sbert_cosine_loss}) is significantly better (empirical evidence from our datasets). Our evaluation was limited to internal datasets due to the lack of public datasets with ad text and metadata (\emph{e.g.}, category, advertiser, and campaign information); exisiting e-commerce datasets with product titles (short, incomplete English sentences) were not suitable for our setup which focuses on native and display ads (with longer text length, complete English sentences).

\paragraph*{Multilingual extension}
Using the multilingual (xlm) version of sentence-BERT\footnote{For training SAS-BERT on ads from Taiwan and Hong Kong, we used distiluse-base-multilingual-cased-v2
in \url{https://www.sbert.net/docs/pretrained_models.html}.
For both Taiwan and Hong Kong ads, we used: positive-to-negative sampling ratio = 1:30, batch size = 30, and epochs = 3.}, we
trained SAS-SBERT models for ads from Taiwan and Hong Kong.
For Taiwan ads, sentence-BERT had a $P@1$ of $0.7$, while SAS-SBERT achieved a $P@1$ of $0.85$ ($21.4\%$ lift). For Hong Kong ads, sentence-BERT had a $P@1$ of $0.83$, while
SAS-SBERT achieved a $P@1$ of $0.94$ ($13.2\%$ lift). Clearly, the SAS-SBERT lifts for Chinese ad text are even stronger compared to their lifts in the case of US based ads in English.
\begin{figure}[]%!htb
\centering
  \includegraphics[width=0.9 \columnwidth]{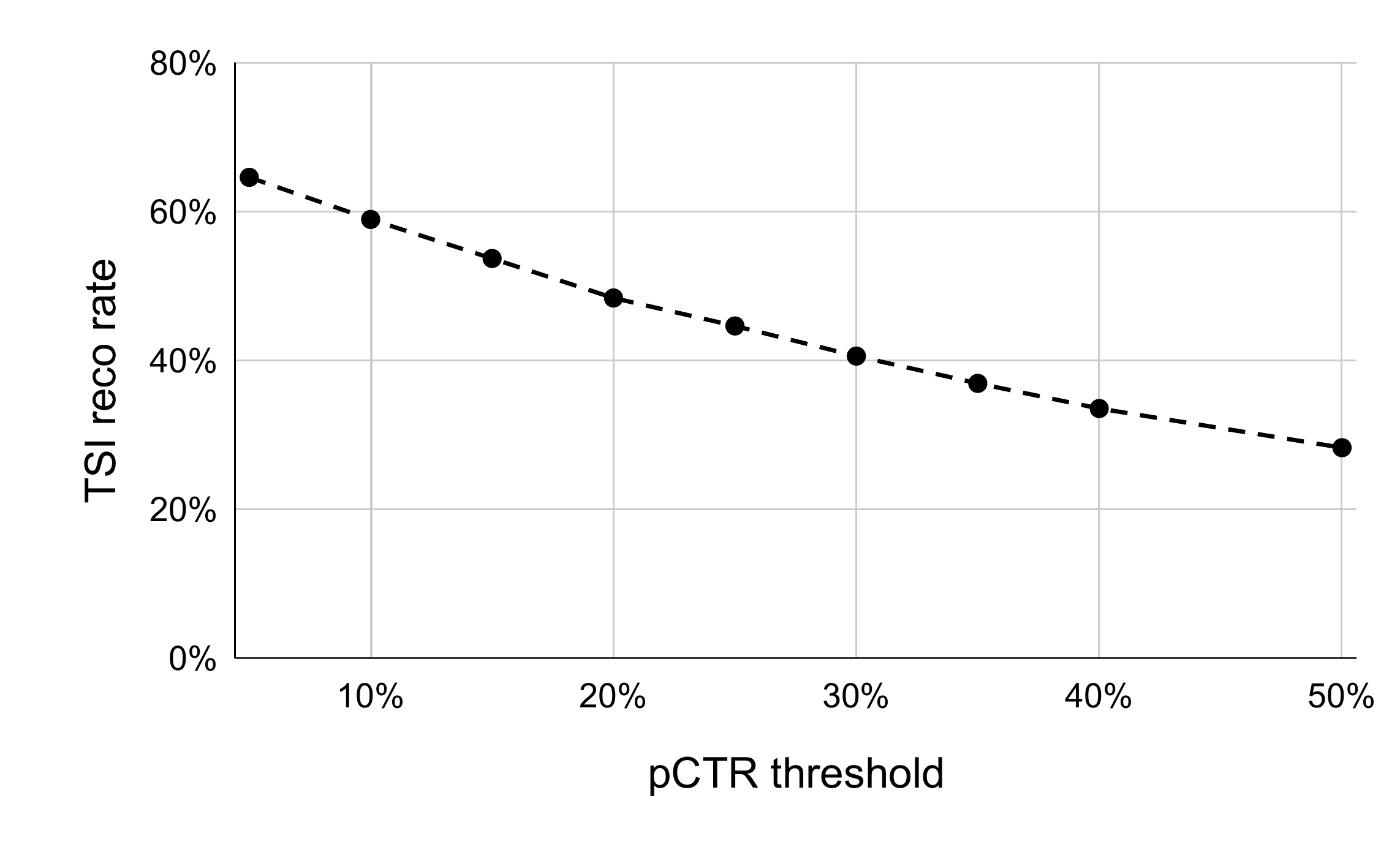}
  \caption{Fraction of ads eligible for recommendations (TSI=0) versus the relative pCTR difference threshold $\Delta$.
  %\yifan{the x and y-labels  are too small}
  }
  \label{fig:pctr_threshold_tuning}
\end{figure}
\subsection{TSI recommendation rate (offline)} \label{sec:tsi_offline_reco_rate}
For TSI, the relative pCTR difference threshold $\Delta$ (as defined in Section~\ref{sec:tsi}) is an important tuning parameter deciding the fraction of ads eligible for recommendations. To get an offline estimate of the fraction of ads eligible for recommendations (\emph{i.e.}, TSI$=0$), we computed TSI for each ad in the test set ($80-20\%$ train-test split), considering the train set as the existing pool of ads. In other words, we fetched top $k=5$ similar ads from the pool of train ads, and computed TSI by comparing their pCTRs with the test ad's pCTR. For similar ads retrieval, the minimum cosine similarity threshold was set to $0.6$.
Figure~\ref{fig:pctr_threshold_tuning} shows the variation of recommendation rate (fraction of ads shown recommendations) with respect to the relative pCTR difference threshold. As $\Delta$ increases, there are fewer ads in a given input's neighborhood with relative pCTR  above the threshold; hence the chances of being labeled weak (TSI=0) decreases.
At $\Delta=20\%$, $\sim 50\%$ of test ads are eligible for recommendations; in other words, $50\%$ of test ads are such that they have at least one neighbor who has $20\%$ better pCTR.
\begin{figure*}[]%!htb
\centering
  \includegraphics[width=2.0 \columnwidth]{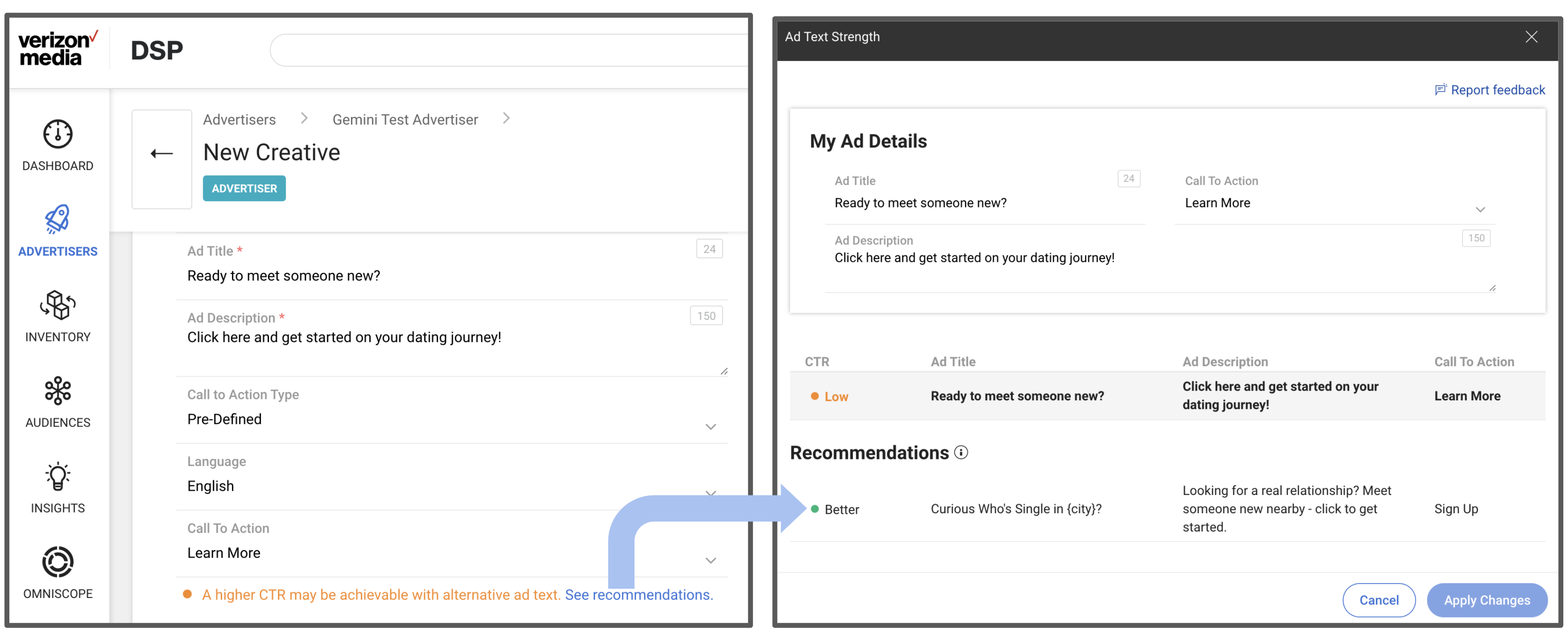}
  \caption{The TSI feature in action for an advertiser in the Verizon Media (DSP) ad platform. The advertiser is given a chance to click `see recommendations' while creating the ad, and clicking it shows the TSI window with suggestions. As shown in the example, mentioning the city, and changing call-to-action to `sign up' can lead to a better CTR for a dating ad.}
  \label{fig:tsi_binary_screen_shot}
\end{figure*}
\subsection{TSI online results}
Our online results for TSI can be categorized into two parts:
(i) aggregated feedback from internal account teams and creative strategists at Verizon Media, and (ii) session-wise log analysis of advertiser (external) interactions with the TSI feature in the Verizon Media (DSP) ad platform. For online testing, the proposed BERT based CTR predictor was used, whereas for similar ads retrieval the word2vec version (as described in Section~\ref{sec:similar_ads_retrieval_results}) was used due to latency constraints. End-to-end online latency for the TSI feature was below 1 second, giving the user real time feedback with their ad text.

\subsubsection{Account team feedback}
We deployed an internal version of TSI for Verizon Media account teams and creative strategists, where they had access to the pCTR scores, all similar ads (and not just suggestions with better pCTR), and final TSI scores with recommendations. The aggregate rating collected from $15$ account strategists on the feature as a whole was $4.1/5$. This was based on $\sim 800$ unique ad texts tried by the account teams (collectively) over a span of 2 months.
Qualitative feedback revolved around TSI being fast, and proving to be a helpful tool. Critical feedback included the usage of conversion prediction for differentiating between click-bait text and genuinely well written ad text, and the detection of subtle yet abusive content which may have high CTRs logged in data -- both are directions for future research for the authors.

%Qualitative feedback was around incorporating abusive language detection, 

\subsubsection{Advertiser session log analysis}
Verizon Media (Yahoo) DSP ad platform is a major ad platform in the online advertising industry. For online tests with real advertisers, we deployed the TSI feature to assist onboarding Verizon Media DSP advertisers.
Figure \ref{fig:tsi_binary_screen_shot} shows a screen shot of the DSP TSI feature for a dating ad example.

\begin{figure}[]%!htb
\centering
  \includegraphics[width=1.0 \columnwidth]{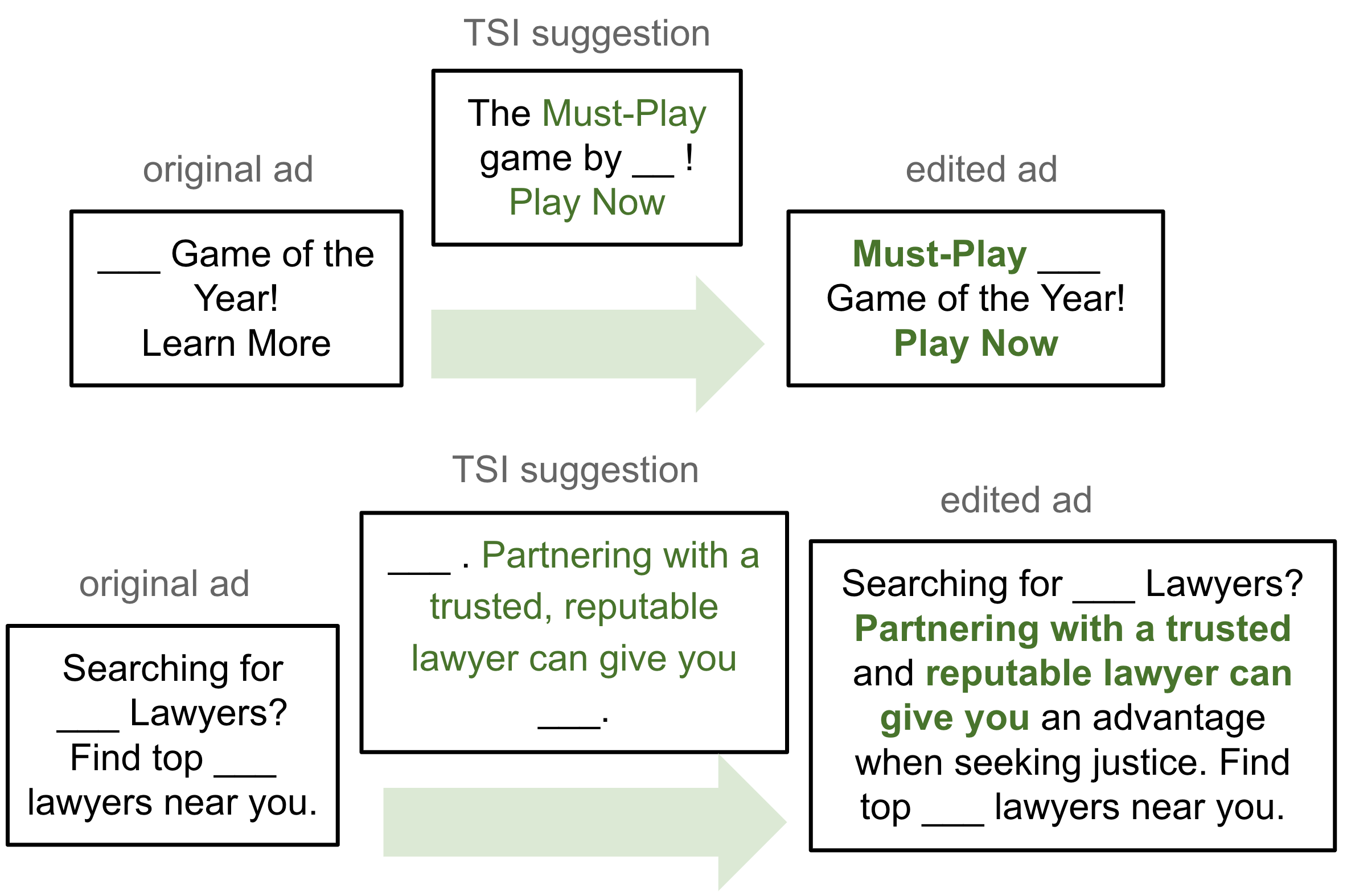}
  \caption{Two anonymized examples from advertiser logs showing the positive impact of TSI. In the top example, `Must-Play' and `Play Now' were added after seeing the TSI suggestion. In the bottom example, the advertiser emphasized on trust and reputation in the edited ad for a law firm.}
  \label{fig:tsi_real_examples}
\end{figure}

We collected sampled data spanning over $1100$ advertiser sessions with English ads (from mid-January to mid-February 2021); a session being a continuous series of online events in the DSP UI (involving the same advertiser) with time gaps no more than 30 minutes between consecutive events.
The following statistics were obtained from the above sessions.
\begin{itemize}
    \item In $56\%$ of sessions, at least one of the ad text variations got a TSI score of $0$, and the advertiser was shown recommendations. This is reasonably close to the offline recommendation rate estimate of $50\%$ with the pCTR difference threshold $\Delta=20\%$ as explained in Section~\ref{sec:tsi_offline_reco_rate}.
    \item In $25.12\%$ of the sessions with TSI recommendations, after seeing the TSI recommendations, the advertiser incorporated a new word (excluding stop-words in NLTK \cite{nltk}) in their ad text. This new word was not present before in the original ad text, but was borrowed from the TSI recommendations by the advertiser. This is a proxy to automatically gauge the influence of the recommendation on the ad text. Furthermore, the $25.12\%$ fraction of advertisers who adopted recommendations, collectively saw a $27\%$ CTR lift over advertisers who were exposed to recommendations but did not adopt.
\end{itemize}
Since the advertisers eventually go ahead with one single ad text,
it is hard to compare the CTRs of the original ad and the final ad after interacting with TSI (the $27\%$ CTR lift reported above is computed by treating adopters and non-adopters separately, but the comparison is not ideal). The ideal way to set up the comparison would be to integrate TSI with a dynamic creative optimization (DCO) framework; this is beyond the scope of this paper (but is a part of our on-going and future work).
To get a qualitative perspective, we manually analyzed a few sessions to find examples where TSI convinced the advertiser towards making positive changes. We have described below two such examples (with anonymization as applicable) in Figure~\ref{fig:tsi_real_examples}. In the top example in Figure~\ref{fig:tsi_real_examples}, a gaming advertiser noticed a TSI suggestion (another successful gaming ad) which highlighted `Must-Play' and had `Play Now' as the call-to-action. Subsequently, the advertiser changed the original ad to incorporate the above changes (as shown in Figure~\ref{fig:tsi_real_examples}), and proceeded with a strong TSI score. In the bottom example, the advertiser introduced words on trust and reputation in the ad for lawyers. As mentioned above, $25.12\%$ of advertiser sessions which saw TSI suggestions, incorporated changes; this demonstrates the significant impact of TSI on advertisers, and Figure~\ref{fig:tsi_real_examples} just shows two such examples. It should also be noted that the TSI suggestions are essentially anonymized versions of older ads with significant number of impressions and clicks in the past, \emph{i.e.}, ads which are already public in nature (as opposed to learning from ads which have not been exposed to the public yet). 
%\yifan{Thee two examples are  difficult to read. Perhaps put in Figured and prettify?}

\section{Discussion} \label{sec:discussion}
Our online results demonstrate the efficacy of TSI in detecting ad text likely to have (relatively) low CTR, and inspiring advertisers to make positive changes in their ad text. We are continuously collecting data from TSI (still an actively used feature in the Verizon Media DSP UI at the time of this submission), and our next steps include using such data for training text generation models along the lines of \cite{cikm2020_createbetterads} to minimize the effort needed from the advertiser's side.
%Our next steps include leveraging retrieval augmented generation models to generate refined ad text based on high pCTR similar ads in the semantic neighborhood.
Increasing our ads data pool to multiple months while accounting for CTR seasonality, is another direction for future research.

%\section*{Acknowledgement}
%The authors want to thank Avinash Chukka (for product guidance, and the ask for the title strength feature), Ryan Hill and Karthikeyan M. for guidance on ad.com use cases. 
\begin{comment}
A notable effort in the process of creating the backend went in for (BERT) model export. Since the model comes as $tf.estimator$, a framework whose $model.predict$ function is not suitable for long-term serving. It loads weights from disk on every prediction request, and will cause the WSGI workers to crash. As a solution we export the model to SavedModel format, which can either be loaded into Python as an object, or to be loaded by TensorFlow Serving's Docker image. In our case, we keep everything in Python since BERT's input requires tokenization, which can't be loaded by the Docker image.
\end{comment}
\bibliographystyle{ACM-Reference-Format}
\bibliography{refs}
%\newpage
%\appendix
%\input{yoca_kdd2021/sections/reproducibility}
\end{document}